\documentclass[letterpaper]{article} 
\usepackage[preprint]{aaai2027}  
\usepackage[hyphens]{url}  
\usepackage{graphicx} 
\usepackage{natbib}  
\usepackage{caption} 
\usepackage{algorithm}
\usepackage{algorithmic}
\usepackage{amsmath} 
\usepackage{multirow}
\usepackage{makecell}
\usepackage{amssymb}
\usepackage{newfloat}
\usepackage{listings}
\DeclareCaptionStyle{ruled}{labelfont=normalfont,labelsep=colon,strut=off} 
\floatstyle{ruled}
\newfloat{listing}{tb}{lst}{}
\floatname{listing}{Listing}

\usepackage{booktabs}

\usepackage{tcolorbox}

\title{OBC-Prune: Outcome-Based Calibration for Large Reasoning Model Pruning}
\author {
    Ha Lan Nguyen\textsuperscript{\rm 1},
    Huy Hoang Tran\textsuperscript{\rm 1},
    Trac-Duy Tran\textsuperscript{\rm 1}\textsuperscript{\rm 2}\equalcontrib\corresponding, 
    Dung D. Le\textsuperscript{\rm 1}\equalcontrib\corresponding
}
\affiliations {
    \textsuperscript{\rm 1}VinUniversity\\
    \textsuperscript{\rm 2}Johns Hopkins University\\
    {lan.nth, hoang.th2, dung.ld}@vinuni.edu.vn,
    trac@jhu.edu
    
}

\begin{document}

\maketitle

\begin{abstract}
Large reasoning models (LRMs) generate long chain-of-thought traces before answering, creating significant inference overhead. Pruning can reduce this cost, but its effectiveness depends on the calibration data used to estimate parameter importance. Recent work calibrates on the model's own rollouts instead of generic dataset, but treats all reasoning tokens uniformly, regardless of whether they contribute to successful reasoning. As a result, pruning protects weights by statistical salience rather than by their contribution to correct reasoning, so weights behind erroneous computation survive as readily as those behind correct computation. These erroneous patterns then get carried into the pruned model, degrading reasoning quality, producing both lower accuracy and longer reasoning traces. We propose Outcome-Based Calibration for Large Reasoning Model Pruning (OBC-Prune) to close this gap. OBC first constructs difficulty-matched pairs of correct and incorrect rollouts from problems the model answers inconsistently. It then estimates the causal importance of each reasoning sentence through intervention-based analysis, quantifying how removing its influence affects subsequent predictions. These causal importance scores are converted into per-token weights that rescale the calibration activations used by one-shot pruning methods (SparseGPT, Wanda, ALPS), without modifying the underlying pruning algorithms. Experiments on DeepSeek-R1-Distill-Qwen 1.5B, 7B, and 14B models at 40\% and 50\% sparsity demonstrate consistent improvements over state-of-the-art calibration baselines across most model sizes and sparsity levels on MATH500, LiveCodeBench, and AIME 2025. These results indicate that preserving causally important reasoning circuits is a substantially more effective pruning objective than uniformly preserving observed activations.

\end{abstract}


\section{Introduction}
\label{sec:intro}

Large reasoning models (LRMs) such as the DeepSeek-R1 series achieve strong performance on challenging reasoning benchmarks by using large-scale reinforcement learning to work through an extended chain of thought (CoT) before committing to a final answer \cite{deepseek_r1}. This extended deliberation is central to their accuracy gains, but it also ties inference cost directly to how long the model chooses to think, making LRMs substantially more resource-intensive to deploy at scale than a conventional instruction-following LLM of comparable size.
 
The inference overhead has fueled broad interest in LLM compression. Model pruning \cite{NIPS1989_6c9882bb, NIPS1992_303ed4c6}, which removes redundant parameters while preserving task performance, has emerged as a popular technique for cutting model latency and computational cost. Most existing pruning methods estimate which parameters are safe to remove from a small amount of data, referred to as calibration data, and this has proven highly effective for conventional LLMs \cite{sparsegpt, wanda, alps}. Applying the same methods to reasoning models, however, causes accuracy to drop far more sharply than it does for conventional LLMs at the same sparsity. One potential cause is a calibration mismatch: these methods estimate weight importance from calibration activations, and generic corpora such as C4~\cite{Raffel2019ExploringTLc4} never activate the reasoning-specific circuits that emerge during RL post-training \cite{emergent_reasoning_heads}, causing standard pruning metrics to misjudge critical parameters.

Reasoning-Aware Compression \cite{rac} and Selective Self-Generated Reasoning \cite{ssgr} address this mismatch by calibrating instead on chain-of-thought rollouts the model generates on its own, recovering much of the accuracy that generic calibration data loses. Both approaches, however, treat every token in a rollout as equally informative regardless of whether that rollout ultimately reached a correct answer. Because the resulting importance scores are derived purely from activation and weight magnitudes rather than answer correctness, weights behind erroneous computation can be preserved as readily as those behind correct computation, which we later show in Section \ref{sec:experiments}: pruned models drift further from the dense model's internal computation, and their chain-of-thought is more likely to run on without ever terminating.
 
We introduce \textbf{OBC-Prune} (Outcome-Based Calibration for Large Reasoning Model Pruning) to close this gap. Rather than treating every calibration token identically, OBC-Prune builds difficulty-matched pairs of correct and incorrect CoT rollouts, scores each reasoning sentence's causal contribution to the final outcome via an attention-suppression intervention, and converts these scores into per-token weights that upweight the activations behind correct rollouts and downweight those behind incorrect ones, all without modifying the underlying pruning solver.

In this work we focus specifically on one-shot pruning \cite{sparsegpt, wanda, alps} of LRMs, compressing the model without any subsequent retraining. Although the DeepSeek-R1-distilled models are released with open weights, the full training and distillation pipeline needed to recover their accuracy is not entirely public \cite{huggingface_openr1}, and retraining a model at this scale is in any case expensive, typically demanding a cluster of several GPUs rather than the single GPU that one-shot pruning typically requires.
 
Our contributions are: (1) a calibration framework that stratifies LRM calibration data by outcome under difficulty-matched pair construction; (2) an interventional, attention-suppression-based causal importance score for reasoning sentences that requires no cross-sequence alignment; and (3) a per-token re-weighting scheme that integrates these scores into the calibration statistics of a one-shot pruning method without modifying its solver, which is demonstrated on three structurally different backends (SparseGPT, Wanda, and ALPS) with markedly different sensitivity to calibration quality.

\section{Preliminaries}
\label{sec:preliminaries}
\paragraph{Large reasoning models.} A core difference between a large reasoning model (LRM) and a conventional instruction-tuned large language model (LLM) is what it is trained to produce: given a prompt $x$, it generates a chain-of-thought $c = (c_1, \ldots, c_T)$ before its final answer $y$, i.e. $(c, y) \sim \pi_\theta(\cdot \mid x)$, rather than answering directly. This behavior is instilled by large-scale reinforcement learning against a verifiable outcome reward (i.e. exact-match on a final numeric answer, unit-test passes for code), which rewards $y$ being correct but places no explicit penalty on the length of $c$, popularized by DeepSeek-R1's use of Group-Relative Policy Optimization \cite{deepseek_r1}. Because longer deliberation is empirically associated with higher accuracy, the length of generated chain-of-thought $T$ often reaches an order of magnitude beyond the prompt length $|x|$. Since serving cost and latency scale with the number of tokens generated, this decode-dominated computation is what makes LRMs substantially more expensive to deploy at scale than a conventional LLM of comparable size.
 
\paragraph{LLM pruning methods.} Most modern language models contain billions of parameters, making pruning a key technique for cutting their memory footprint, inference latency, and energy cost without giving up much accuracy. A widely adopted formulation casts this as a per-layer reconstruction problem \cite{obc}: given a dense layer weight $\mathbf{W}_\ell$ and its calibration input activations $\mathbf{X}_\ell$, pruning seeks a weight matrix $\hat{\mathbf{W}}_\ell$ that reproduces the layer's output while satisfying a sparsity budget $S$:
\begin{equation}
  \min_{\hat{\mathbf{W}}_\ell} \left\| \mathbf{W}_\ell \mathbf{X}_\ell - \hat{\mathbf{W}}_\ell \mathbf{X}_\ell \right\|_2^2
  \quad \text{s.t.} \quad \left\| \hat{\mathbf{W}}_\ell \right\|_0 \le S.
  \label{eq:pruning-objective}
\end{equation}
Unlike structured pruning \cite{llm_pruner, sheared_llama} and training-integrated pruning \cite{movement_pruning}, both of which require post-pruning fine-tuning or repeated forward-backward passes to recover accuracy, one-shot (post-training) pruning estimates $\hat{\mathbf{W}}_\ell$ directly from Eq.~\ref{eq:pruning-objective} using only a small calibration set and a single forward pass, without any retraining.

Within one-shot pruning, SparseGPT \cite{sparsegpt} solves Eq.~\ref{eq:pruning-objective} via the layer's Hessian $\mathbf{H}_\ell = \mathbf{X}_\ell \mathbf{X}_\ell^\top$, pruning weight $q$ by the Optimal Brain Surgeon saliency $w_q^2 / (\mathbf{H}_\ell^{-1})_{qq}$ and updating surviving weights to compensate, so the calibration set governs which weights survive through the full curvature of $\mathbf{H}_\ell$. Wanda \cite{wanda} uses the simpler score $s_{ij} = |w_{ij}| \cdot \|\mathbf{x}_j\|_2$, a diagonal approximation with the same calibration dependence but no compensation update. ALPS \cite{alps} instead solves Eq.~\ref{eq:pruning-objective} via operator splitting (ADMM) followed by a conjugate-gradient polishing step rather than greedy OBS updates, but like SparseGPT is driven by the Gram matrix $\mathbf{X}_\ell\mathbf{X}_\ell^\top$. We adopt these three structurally distinct solvers to test how far OBC-Prune's calibration re-weighting generalizes.

\section{Related Work}
\label{sec:related-work}

\paragraph{Compression and calibration for reasoning models.}
Post-training compression, such as weight pruning \cite{sparsegpt, wanda, alps} and quantization \cite{frantar2023gptqaccurateposttrainingquantization, MLSYS2024_awq}, is the standard route to reducing LLM inference cost without retraining. Applied to LRMs, however, both degrade accuracy disproportionately relative to conventional LLMs of comparable size \cite{zhang2026when, lotfi2026quantizedreasoningmodelsthink}, indicating that the circuits supporting extended reasoning are unusually fragile under compression.
One source of this fragility is a calibration mismatch: generic corpora such as C4 are dominated by prompt-like text, whereas an LRM's chain of thought is far longer than its prompt and is produced by reasoning-specific circuits that generic text never exercises \cite{emergent_reasoning_heads}, so calibrating on C4 estimates weight importance almost entirely from computation the model never performs while reasoning.
In the context of post-training pruning for LLMs, the selection of calibration data has been shown to have an impact comparable to that of the pruning algorithm \cite{beware_calibration}. Because the original pretraining distribution is typically inaccessible, this motivates to use data that the model generates on its own as a proxy for it. RAC \cite{rac} and SSGR \cite{ssgr} extend this principle to reasoning models specifically, calibrating instead on the model's own on-policy CoT rollouts and recovering much of the accuracy lost under generic calibration; SSGR further filters rollouts by difficulty and trace length, showing that RAC's undifferentiated use of all on-policy traces is itself suboptimal. The same reasoning-specific mismatch extends beyond one-shot weight pruning: in structured pruning, RESP \cite{resp} replaces activation reconstruction with decode-only gradient importance computed from self-generated calibration traces, and \citet{llms_to_lrms} show that reasoning-augmented models require calibration and post-pruning recovery data matched to their own training distribution, with the best-performing structured-pruning strategy itself differing between instruction-following and reasoning-augmented models. Complementary mechanistic work identifies reasoning-specific attention heads that generic calibration methods never activate \cite{emergent_reasoning_heads}.
None of these approaches stratify calibration by whether the underlying rollout reached a correct answer, nor use an interventional signal to identify which reasoning steps causally drove that outcome. OBC-Prune fills this gap: it uses continuous per-token weights derived from a measured causal quantity, over a calibration set of difficulty-matched correct/incorrect pairs - a construction borrowed from preference-pair mining \cite{rafailov2024directpreferenceoptimizationlanguage}, used here to shape calibration statistics rather than a training objective.

\paragraph{Attributing outcomes to reasoning steps.} Causal tracing \cite{causal_tracing} and path patching \cite{path_patching} identify which hidden states or computational paths causally drive a behavior, via activation corruption/restoration or path tracing between components. Thought Anchors \cite{thought_anchors} applies the same interventional logic to CoT reasoning, suppressing a sentence's attention contribution to later tokens and measuring the resulting shift in predictions to identify which sentences drive a rollout's trajectory.
A parallel line of work scores reasoning steps externally: process reward models are trained on human \cite{lightman2024letsverify} or Monte-Carlo-estimated \cite{wang2024mathshepherdverifyreinforcellms} step-level labels to judge whether each step is correct. Compared to these works, the interventional measurement is training-free and requires no step-level supervision, transfers across domains without a domain-specific reward model, and measures the model's own downstream dependence on a sentence rather than an external judge's verdict on that sentence's correctness.
OBC-Prune adopts this attention-suppression measurement as its sentence-level causal score and, to our knowledge, is the first to route such a score into a pruning-calibration objective rather than using it for post-hoc interpretation.

\section{Methodology}
\label{sec:methodology}
 OBC-Prune calibrates a one-shot pruning backend on outcome-based chain-of-thought rollouts. As summarized in Algorithm~\ref{alg:OBC}, it proceeds in four stages: (1) \emph{matched pair construction} samples a correct and an incorrect rollout per qualifying problem; (2) \emph{attention-suppression scoring} measures each sentence's causal contribution to the rollout's outcome; (3) convert  into per-token weights that upweight correct-rollout activations and downweight incorrect-rollout activations; and (4) these weights rescale calibration activations before they enter the pruning backend's second-order statistic, without modifying its solver.

\begin{algorithm}[t]
\caption{OBC-Prune}
\label{alg:OBC}
\textbf{Input:} $\mathcal{P}$: problem pool; $f_\theta$: dense LRM with $L$ layers; $n$: rollouts per problem; $N$: calibration budget; $s$: target sparsity \\
\textbf{Output:} $\hat{f}$: compressed model
\begin{algorithmic}[1]
\STATE \textbf{Stage 1: Matched pair construction}
\FOR{$p \in \mathcal{P}$, until $|\mathcal{P}'|=N$}
    \STATE sample $\{R_i\}_{i=1}^n \sim f_\theta(\cdot\mid p)$ at temperature $\tau{=}0.8$; skip $p$ unless outcomes contain both correct and incorrect rollouts
    \STATE $(c_p, w_p) \gets$ median-length correct, longest near-miss incorrect rollout; $\mathcal{P}' \gets \mathcal{P}' \cup \{(c_p,w_p)\}$
\ENDFOR
\STATE \textbf{Stage 2: Attention-suppression scoring}
\FOR{$(c_p, w_p) \in P'$, $R \in \{c_p, w_p\}$}
    \STATE segment $R$ into sentences $\{S_k\}_{k=1}^M$; compute $\text{effect}(k, R)$ via (2)--(4) $\forall k$
    \STATE $\tilde{e}(k, R) \leftarrow \text{softmax}_k\big(\text{effect}(k, R)\big)$
\ENDFOR
\STATE \textbf{Stage 3: Per-token weight conversion}
\FOR{$(c_p, w_p) \in P'$}
    \STATE for $t \in c_p$: $w_t^{\text{correct}} \leftarrow 1 + \beta \cdot \tilde{e}(\text{seg}(t), c_p)$ \quad via (6)
    \STATE for $t \in w_p$: $w_t^{\text{wrong}} \leftarrow \max\big(\varepsilon,\, 1 - \gamma \cdot \tilde{e}(\text{seg}(t), w_p)\big)$ \quad via (7)
\ENDFOR
\STATE \textbf{Stage 4: Calibration-statistic re-weighting}
\STATE $\tilde{C}_\ell \leftarrow 0 \ \forall \ell$
\FOR{$(c_p, w_p) \in P'$}
    \STATE accumulate $\tilde{C}_\ell \mathrel{+}= \sum_{t \in c_p} w_t^{\text{correct}} \, x_t^\ell (x_t^\ell)^\top$ and $\tilde{C}_\ell \mathrel{+}= \sum_{t \in w_p} w_t^{\text{wrong}} \, x_t^\ell (x_t^\ell)^\top$, $\ \forall \ell$ \quad via (8)
\ENDFOR
\FOR{$\ell = 1, \dots, L$}
    \STATE $\hat{W}_\ell \gets \textsc{Prune}(W_\ell, \tilde{C}_\ell, s)$ 
\ENDFOR
\RETURN $\hat{f} = \{\hat{W}_1,\dots,\hat{W}_L\}$
\end{algorithmic}
\end{algorithm}
\subsection{Stage 1: Matched Pair Construction}
 
For each candidate problem $p$, we sample $n{=}16$ independent chain-of-thought rollouts
from the dense model (temperature $0.8$) and retain $p$ only if at least one rollout
reaches a correct final answer and at least one reaches an incorrect one; problems on
which all 16 rollouts agree (always correct or always incorrect) are discarded, since no
matched correct/wrong pair exists to isolate the outcome effect. We sample problems and
apply this filter until $N{=}128$ qualifying problems have been collected, fixed in advance
as our calibration budget.
For each qualifying problem, we collect the median-length correct rollout $c_p$ and the
longest \emph{near-miss} wrong rollout $w_p$ (complete wrong answer, length $\ge 50\%$ of
median correct length). This reflects the two rollouts' different roles: $c_p$
contributes calibration mass to the reconstruction statistic itself, where prior evidence
favors moderate length over either extreme \cite{ssgr}, motivating the median-length
choice; by contrast, $w_p$ is selected to maximize the resolution of the causal scoring in Section~\ref{sec:stage2}, where the achievable resolution for localizing exactly where the reasoning diverged toward failure is bounded by the number of sentence spans available to score. We therefore select the longest available near-miss
wrong rollout to maximize the number of causally scoreable candidate sentences, rather
than optimizing the length of $w_p$ for reconstruction quality as we do for $c_p$.
The matched-pair design ensures that activation differences between $c_p$ and $w_p$
isolate the outcome effect and remove the difficulty confound, since both rollouts
originate from the very same problem.
 
\subsection{Stage 2: Attention Suppression Scoring}
 \label{sec:stage2}
In this stage, each sentence is assigned a weight reflecting its causal importance to the
rollout's outcome via an interventional measure that requires no cross-sequence alignment.

\paragraph{Sentence segmentation and base pass.}
Each rollout $R \in \{c_p, w_p\}$ is segmented into sentence spans
$\{S_k = [a_k, b_k]\}_{k=1}^{M}$ by paragraph boundary.
A single full forward pass over $R$ yields base logits $L_t \in \mathbb{R}^{|V|}$
and next-token distributions $p_t = \mathrm{softmax}(L_t)$, and caches the key-value
states for all layers.

\paragraph{Per-sentence causal effect.}
For each sentence $S_k$, we construct a suppressed attention mask
\begin{equation}
  \tilde{A}[t,t'] = A[t,t'] \cdot \bigl(1 - \mathbf{1}[a_k \le t' \le b_k, \, t > b_k]\bigr),
\end{equation}
which prevents all future tokens $t > b_k$ from attending to $S_k$ while leaving the
causal structure elsewhere intact.
Re-running the forward pass from position $a_k$ under $\tilde{A}$ yields suppressed
logits $\tilde{L}_t^{(k)}$ and suppressed next-token distributions
$\tilde{p}_t^{(k)} = \mathrm{softmax}(\tilde{L}_t^{(k)})$ for all $t > b_k$.
We define the raw causal effect as
\begin{equation}
  \mathrm{effect}_{\mathrm{raw}}(k,R)
  =\frac{1}{T-b_k}\sum_{t=b_k+1}^{T}
  \mathrm{KL}\!\left(p_t\,\|\,\tilde{p}^{(k)}_t\right).
\end{equation}
Earlier spans can affect more downstream positions by construction \cite{thought_anchors}. We mitigate this positional advantage by centering each score against the cumulative effects of earlier spans
\begin{equation}
  \mathrm{effect}(k,R)=\mathrm{effect}_{\mathrm{raw}}(k,R)
  -\frac{1}{k-1}\sum_{j=1}^{k-1}\mathrm{effect}_{\mathrm{raw}}(j,R),
\end{equation}

where the first sentence has no earlier sentences to correct against and is scored by its raw effect alone; for $k>1$, the subtracted term removes the advantage that earlier sentences have from
causally influencing more downstream tokens by construction.
Per-rollout sentence weights are then obtained via softmax
\begin{equation}
  \tilde{e}(k,R) = \frac{\exp(\mathrm{effect}(k,R))}{\sum_{k'=1}^{M}\exp(\mathrm{effect}(k',R))}.
\end{equation}
 
\subsection{Stage 3: Per-Token Weight Conversion}
Section \ref{sec:stage2} yields a causal-importance score for each sentence, but the calibration
statistic consumed by one-shot pruning solvers is accumulated over
individual tokens rather than sentences. This stage converts the sentence-level
scores $\tilde{e}(k, R)$ from Section \ref{sec:stage2} into a per-token weight,
upweighting tokens drawn from causally important sentences of the correct rollout
and downweighting those from causally important sentences of the wrong rollout,
before using these weights to rescale calibration activations in Section \ref{sec:stage4}.

For each token $t$ let $\mathrm{seg}(t) = k$ denote the index of the sentence span
$S_k=[a_k,b_k]$ containing $t$.
Tokens from correct rollouts are upweighted by their sentence's causal importance;
tokens from wrong rollouts are downweighted:
\begin{equation}
  w_t^{\mathrm{correct}} = 1 + \beta \cdot \tilde{e}(\mathrm{seg}(t),\, c_p),
\end{equation}
\begin{equation}
  w_t^{\mathrm{wrong}} = \max\!\left(\varepsilon,\; 1 - \gamma \cdot \tilde{e}(\mathrm{seg}(t),\, w_p)\right),
\end{equation}
where $\beta > 0$ controls upweighting strength, $\gamma \in [0,1]$ controls
downweighting strength, and $\varepsilon = 0.1$ prevents full suppression of
wrong-rollout tokens to keep the Hessian positive definite.

\subsection{Stage 4: Calibration-Statistic Re-Weighting}
 \label{sec:stage4}
The per-token weights $w_t^{\mathrm{correct}}$ and $w_t^{\mathrm{wrong}}$ are injected purely
by rescaling each calibration token's activation vector $x_t^\ell$ by $\sqrt{w_t}$
\emph{before} it enters whatever second-order statistic a one-shot pruning backend
accumulates from calibration data, rather than by modifying the backend's solver.
Standard one-shot pruning accumulates this statistic uniformly over calibration tokens;
OBC-Prune instead accumulates the outcome-weighted statistic
\begin{equation}
  \tilde{\mathbf{C}}_\ell = \sum_p\!\left[
    \sum_{t \in c_p} w_t^{\mathrm{correct}} \, x_t^\ell (x_t^\ell)^\top
    + \sum_{t \in w_p} w_t^{\mathrm{wrong}} \, x_t^\ell (x_t^\ell)^\top
  \right].
\end{equation}
This single rescaling operation re-weights all three backends we consider, each of which
consumes a different second-order statistic $\mathbf{C}_\ell$ built from the same calibration
activations $\mathbf{X}_\ell$. For SparseGPT \cite{sparsegpt}, which accumulates the layer
Hessian $\mathbf{H}_\ell = \mathbf{X}_\ell \mathbf{X}_\ell^\top$, rescaling gives
$\tilde{\mathbf{H}}_\ell = \mathbf{X}_\ell \, \mathrm{diag}(w) \, \mathbf{X}_\ell^\top$ --
exactly the Hessian of a \emph{weighted} least-squares reconstruction objective, so a
token from a causally important correct-reasoning step incurs a larger reconstruction
penalty if the weights needed to reproduce it are pruned, while one from a causally
important wrong-reasoning step incurs a smaller one. SparseGPT's solver (Cholesky
decomposition, OBS saliency ranking, weight-update rule) is applied unmodified to
$\tilde{\mathbf{H}}_\ell$ in place of $\mathbf{H}_\ell$. Wanda \cite{wanda} and ALPS
\cite{alps} admit the same rescaling under their own second-order statistics, with each
solver likewise left unmodified; we defer this derivation to
Appendix~\ref{sec:backend_reweighting}. In every case, only the curvature or energy
estimate the solver consumes changes, so pruning order now reflects outcome-weighted
rather than uniformly-weighted reasoning activations. Results for all three backends are
reported in Section~\ref{sec:experiments}.
\section{Experiments}
\label{sec:experiments}

In this section, we investigate the following research questions (RQs):

\textbf{RQ1.} Does OBC-Prune improve accuracy over baselines across model scales, sparsity levels, and benchmarks, and do its gains increase with sparsity?

\textbf{RQ2.} Does this improvement generalize to structurally different one-shot pruning backends?

\textbf{RQ3.} Why does OBC-Prune improve accuracy, and is this improvement associated with better preservation of the dense model's computation, leading to fewer reasoning-termination failures and shorter completions?

\textbf{RQ4.} Does OBC-Prune's performance advantage generalize across different language model families and base architectures?

\textbf{RQ5.} Is the improvement robust to hyperparameter choice, and how far does it push the achievable sparsity before both methods collapse?

\begin{table*}[t]
\centering
\setlength{\tabcolsep}{3pt}
\small 
\begin{tabular}{lllcccccccc}
\toprule
\multirow{2}{*}{Benchmark} & \multirow{2}{*}{Model} & \multirow{2}{*}{Sparsity}
& \multicolumn{4}{c}{Accuracy $\pm$ SE}
& \multicolumn{4}{c}{Runtime (min)} \\
\cmidrule(lr){4-7}
\cmidrule(lr){8-11}
& & & C4 & RAC & SSGR & OBC-Prune & C4 & RAC & SSGR & OBC-Prune \\
\midrule
\multirow{9}{*}{MATH500}
& \multirow{3}{*}{1.5B}
& Dense & \multicolumn{4}{c}{0.802 (0.0178)}& \multicolumn{4}{c}{6.8}\\
& & 40\% & 0.692 (0.021) & 0.686 (0.020) & 0.754 (0.019) & \textbf{0.774 (0.019)} & 17.8 & 15.7 & 11.2 & \textbf{10.4}\\
& & 50\% & 0.328 (0.021) & 0.528 (0.022) & 0.700 (0.021) & \textbf{0.706 (0.020)} & 42.8 & 27.8 & 16.6 & \textbf{16.3}\\
\cmidrule(lr){2-11}
& \multirow{3}{*}{7B}
& Dense & \multicolumn{4}{c}{0.886 (0.014)}& \multicolumn{4}{c}{8.5}\\
& & 40\% & 0.840 (0.016) & 0.832 (0.016) & 0.842 (0.016) & \textbf{0.852 (0.016)} & 30.8 & 29.1 & 21.3 & \textbf{20.7}\\
& & 50\% & 0.708 (0.020) & 0.744 (0.0195) & 0.728 (0.020) & \textbf{0.772 (0.018)} & 43.7 & 35.6 & 33.8 & \textbf{27.8}\\
\cmidrule(lr){2-11}
& \multirow{3}{*}{14B}
& Dense & \multicolumn{4}{c}{0.914 (0.014)}& \multicolumn{4}{c}{14.4}\\
& & 40\% & 0.854 (0.016) & 0.870 (0.015) & 0.884 (0.014) & \textbf{0.896 (0.014)} & 26.0 & 21.7 & 22.2 & \textbf{20.9}\\
& & 50\% & 0.788 (0.018) & 0.812 (0.017) & 0.858 (0.016) & \textbf{0.866 (0.015)} & 42.9 & 28.25 & 28.9 & \textbf{23.2}\\
\midrule
\multirow{9}{*}{LiveCodeBench}
& \multirow{3}{*}{1.5B}
& Dense & \multicolumn{4}{c}{0.156 (0.022)}& \multicolumn{4}{c}{20.65}\\
& & 40\% & 0.075 (0.016) & 0.127 (0.020) & 0.075 (0.016) & \textbf{0.130 (0.020)} & 38.4 & 30.4 & 33.8 & \textbf{27.7}\\
& & 50\% & 0.030 (0.010) & 0.067 (0.015) & 0.030 (0.010) & \textbf{0.078 (0.016)} & 44.8 & 60.2 & 39.5 & \textbf{33.8}\\
\cmidrule(lr){2-11}
& \multirow{3}{*}{7B}
& Dense & \multicolumn{4}{c}{0.354 (0.029)}& \multicolumn{4}{c}{26.65}\\
& & 40\% & 0.261 (0.027) & 0.347 (0.029) & 0.302 (0.028) & \textbf{0.352 (0.028)} & 48.7 & 33.8 & 38.5 & \textbf{30.8}\\
& & 50\% & 0.105 (0.019) & 0.281 (0.021) & 0.205 (0.025) & \textbf{0.310 (0.028)} & 86.6 & 34.3 & 59.9 & \textbf{33.1}\\
\cmidrule(lr){2-11}
& \multirow{3}{*}{14B}
& Dense & \multicolumn{4}{c}{0.522 (0.030)}& \multicolumn{4}{c}{75.25}\\
& & 40\% & 0.437 (0.028) & 0.462 (0.030) & 0.463 (0.031) & \textbf{0.485 (0.031)} & 90.0 & 88.3 & 87.4 & \textbf{79.1}\\
& & 50\% & 0.295 (0.028) & 0.417 (0.030) & 0.358 (0.029) & \textbf{0.481 (0.031)} & 183.0 & 123.9 & 156.1 & \textbf{109.8}\\
\midrule
\multirow{9}{*}{AIME-25}
& \multirow{3}{*}{1.5B}
& Dense & \multicolumn{4}{c}{0.166 (0.069)}& \multicolumn{4}{c}{5.31}\\
& & 40\% & 0.100 (0.056) & 0.033 (0.020) & 0.133 (0.063) & \textbf{0.166 (0.069)} & 5.4 & 5.7  & 5.3 & \textbf{4.2} \\
& & 50\% & 0.000 & 0.033 (0.033) & \textbf{0.133 (0.063)} & \textbf{0.133 (0.063)} & 6.5 & 5.8 & 5.4 & \textbf{5.1}\\
\cmidrule(lr){2-11}
& \multirow{3}{*}{7B}
& Dense & \multicolumn{4}{c}{0.33 (0.082)}& \multicolumn{4}{c}{8.41}\\
& & 40\% & 0.30 (0.088) & 0.30 (0.085) & 0.30 (0.091) & \textbf{0.33 (0.085)} & 9.6 & 8.7 & 8.7 & \textbf{8.6}\\
& & 50\% & 0.133 (0.063) & 0.06 (0.046) & \textbf{0.20 (0.074)} & \textbf{0.20 (0.074)} & 11.7 & 9.9 & 9.2 & \textbf{8.7}\\
\cmidrule(lr){2-11}
& \multirow{3}{*}{14B}
& Dense & \multicolumn{4}{c}{0.4 (0.090)}& \multicolumn{4}{c}{17.3}\\
& & 40\% & 0.033 (0.087) & 0.33 (0.087) & \textbf{0.433 (0.092)} & \textbf{0.433 (0.092)} & 25.2 & 22.6 & 23.4 & \textbf{21.2}\\
& & 50\% & 0.233 (0.079) & 0.133 (0.063) & 0.267 (0.082) & \textbf{0.33 (0.090)} & 29.1 & 32.7 & 28.9 & \textbf{28.2}\\
\bottomrule
\end{tabular}%
\caption{Accuracy with standard error (SE) and runtime comparison for dense and sparse DeepSeek-R1-Distill-Qwen models under C4, RAC, SSGR and OBC-Prune calibration using SparseGPT pruning backend, on MATH500~\cite{math500_hf} (acc@1), LiveCodeBench~\cite{livecodebench} (pass@16), and AIME-25~\cite{aime25} (acc@1).}
\label{tab:pruning_results}
\end{table*}

\paragraph{Experimental Setup.} We evaluate OBC-Prune on DeepSeek-R1-Distill-Qwen (1.5B/7B/14B), against pruning using calibration set C4\cite{Raffel2019ExploringTLc4}, RAC \cite{rac}, SSGR \cite{ssgr} as the baselines and the dense model as reference, across three one-shot pruning backends: SparseGPT \cite{sparsegpt}, Wanda \cite{wanda}, and ALPS \cite{alps}, each at 40\% and 50\% sparsity. OBC-Prune's calibration set consists of $N{=}128$ difficulty-matched CoT pairs (each problem sampled 16 times at temperature $0.8$ on OpenR1-Math-220k, retained only if at least one rollout was correct and at least one was incorrect), with $\beta{=}2$, $\gamma{=}0.5$, $\varepsilon{=}0.1$, applied identically across all three backends (Section~\ref{sec:methodology}); both calibration methods run on a single H100 per model. All rollout sampling and generation use a fixed random seed of 42, held constant across model sizes, backends, and calibration methods. We evaluate on MATH500 \cite{math500_hf} (acc@1), LiveCodeBench \cite{livecodebench} (pass@16), and AIME 2025 \cite{aime25} (acc@1), with matched generation settings (temperature $0.6$, top-$p$ $0.95$, max length $32{,}768$, bfloat16) for dense and pruned models, reporting accuracy $\pm$ SE and wall-clock runtime.


\paragraph{Discussion of pruning results (RQ1).} Table~\ref{tab:pruning_results} reports accuracy under SparseGPT across all three benchmarks. OBC-Prune matches or exceeds all baselines at every model size and sparsity level on MATH500 and LiveCodeBench, and its margin over the strongest baseline (SSGR) widens as sparsity increases, most visibly at 7B and 14B under LiveCodeBench/50\% (14B: 0.481 vs.\ 0.358). On AIME-25, where the benchmark's 30-problem size yields wide standard errors, OBC-Prune matches or exceeds SSGR, the strongest baseline. Runtime tracks accuracy: OBC-Prune is the fastest calibration method in nearly every cell, with the advantage most pronounced at higher sparsity and larger models. Taken together, these results support RQ1's premise that OBC-Prune's gains grow with sparsity and hold up against the prior calibration methods.

\begin{table*}[t]
\centering
\small
\begin{tabular}{llcccccccc}
\toprule
\multirow{2}{*}{Model} & \multirow{2}{*}{Sparsity} &
\multicolumn{4}{c}{Accuracy $\pm$ SE} &
\multicolumn{4}{c}{Runtime (min)} \\
\cmidrule(lr){3-6}
\cmidrule(lr){7-10}
& & C4 & RAC & SSGR & OBC-Prune & C4 & RAC & SSGR & OBC-Prune \\
\midrule

\multirow{3}{*}{1.5B}
& Dense & \multicolumn{4}{c}{0.802 (0.018)} & \multicolumn{4}{c}{6.8} \\
& 40\% &
0.606 (0.022) &
0.644 (0.022)&
0.668 (0.021)&
\textbf{0.702 (0.021)}&
26.90&
25.95&
25.60&
\textbf{22.13} \\
& 50\% &
0.378 (0.022) &
0.392 (0.022)&
0.488 (0.022)&
\textbf{0.524 (0.022)}&
45.80 &
46.77&
44.30&
\textbf{41.75}\\

\cmidrule(lr){1-10}

\multirow{3}{*}{7B}
& Dense & \multicolumn{4}{c}{0.886 (0.014)} & \multicolumn{4}{c}{8.5} \\
& 40\% &
0.702 (0.018) &
0.634 (0.022) &
0.775 (0.015)&
\textbf{0.862 (0.015)} &
35.30 &
36.20 &
18.10&
\textbf{14.55}\\
& 50\% &
0.300 (0.021) &
0.430 (0.018)&
0.656 (0.020)&
\textbf{0.716 (0.021)}&
101.80 &
97.05&
51.10&
\textbf{48.87} \\

\cmidrule(lr){1-10}

\multirow{3}{*}{14B}
& Dense & \multicolumn{4}{c}{0.914 (0.014)} & \multicolumn{4}{c}{14.4} \\
& 40\% &
0.858 (0.016) &
0.828 (0.017) &
0.858 (0.014) &
\textbf{0.880 (0.015)} &
24.30 &
33.83 &
25.30 &
\textbf{24.13} \\
& 50\% &
0.772 (0.019) &
0.744 (0.022) &
0.850 (0.016) &
\textbf{0.852 (0.016)} &
47.90 &
49.23 &
42.70 &
\textbf{38.60} \\

\bottomrule
\end{tabular}
\caption{Accuracy with standard error (SE) and runtime comparison for Wanda-pruned DeepSeek-R1-Distill-Qwen models (1.5B/7B/14B) under different calibration datasets (C4, RAC, SSGR, and OBC-Prune) at 40\% and 50\% sparsity on MATH500~\cite{math500_hf} (acc@1). Dense (unpruned) accuracy and evaluation runtime are shown for reference.}
\label{tab:wanda_results}
\end{table*}

\paragraph{Discussion of Wanda and ALPS results (RQ2).} Table~\ref{tab:wanda_results} extends the RQ1 comparison to all four calibration methods under Wanda; the corresponding results for ALPS are given in Appendix~\ref{sec:alps_appendix}. Under Wanda, OBC-Prune is the best-performing method in every model/sparsity cell against all three baselines, with the largest gains at 7B, reaching 0.862 vs.\ RAC's 0.634 at 40\% sparsity and 0.716 vs.\ 0.430 at 50\%. Notably, RAC is not uniformly better than generic C4 under Wanda: at 7B/40\% and 14B/40\%, RAC falls below C4 (0.634 vs.\ 0.702; 0.828 vs.\ 0.858), suggesting on-policy calibration without outcome or difficulty filtering can hurt Wanda more than generic text does. SSGR's filtering recovers most of this gap and is consistently the second-best method, coming close to OBC-Prune at 14B/50\% (0.850 vs.\ 0.852). Runtime tracks the same pattern, with OBC-Prune roughly halving MATH500 evaluation wall-clock over RAC at 7B/50\%. ALPS is far more robust to calibration quality: accuracy improves monotonically from C4 to RAC to SSGR to OBC-Prune in nearly every cell, consistent with Section~\ref{sec:preliminaries}'s observation that ALPS's compensation update is the mechanism Wanda lacks; OBC-Prune's margin over RAC is correspondingly smaller than under Wanda, ranging from a slight edge at 14B/40\% to a substantial gap at 1.5B/50\%.

\begin{figure}[t]
\centering
\includegraphics[width=1.0\linewidth]{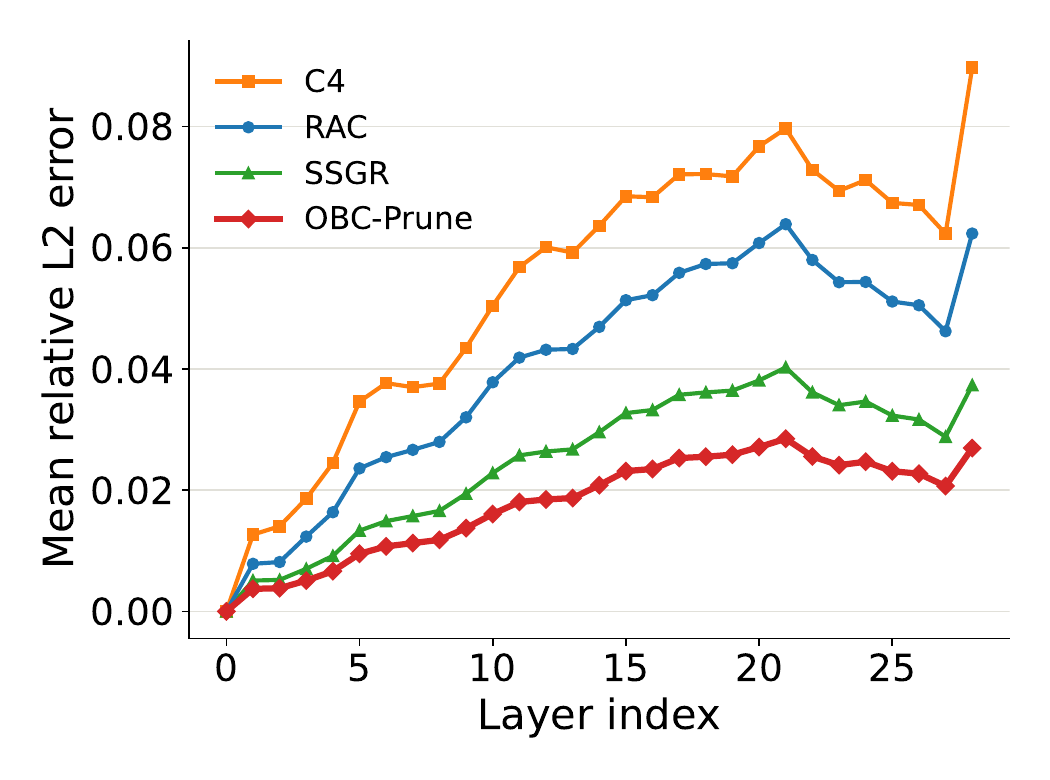}
\caption{Mean relative reconstruction error vs.\ dense hidden states, by layer, for C4, RAC, SSGR, and OBC-Prune}
\label{fig:reconstruction_error}
\end{figure}

\paragraph{Reconstruction fidelity to the dense model (RQ3).} As a complementary, layer-level check, we passed a chain-of-thought through the dense model and through each 40\%-sparsity pruned models, and compared hidden states at every layer via a relative squared error $\|\mathbf{h}_{\text{pruned}} - \mathbf{h}_{\text{dense}}\|_2^2 / \|\mathbf{h}_{\text{dense}}\|_2^2$. Averaged over all layers and token positions, OBC-Prune's hidden states track the dense model's computation more faithfully than those produced under any of the baseline calibration methods (Figure~\ref{fig:reconstruction_error}), consistent with OBC-Prune's calibration objective more closely preserving the dense model's internal computation.

\begin{figure}[h]
\centering
\includegraphics[width=1.0\linewidth]{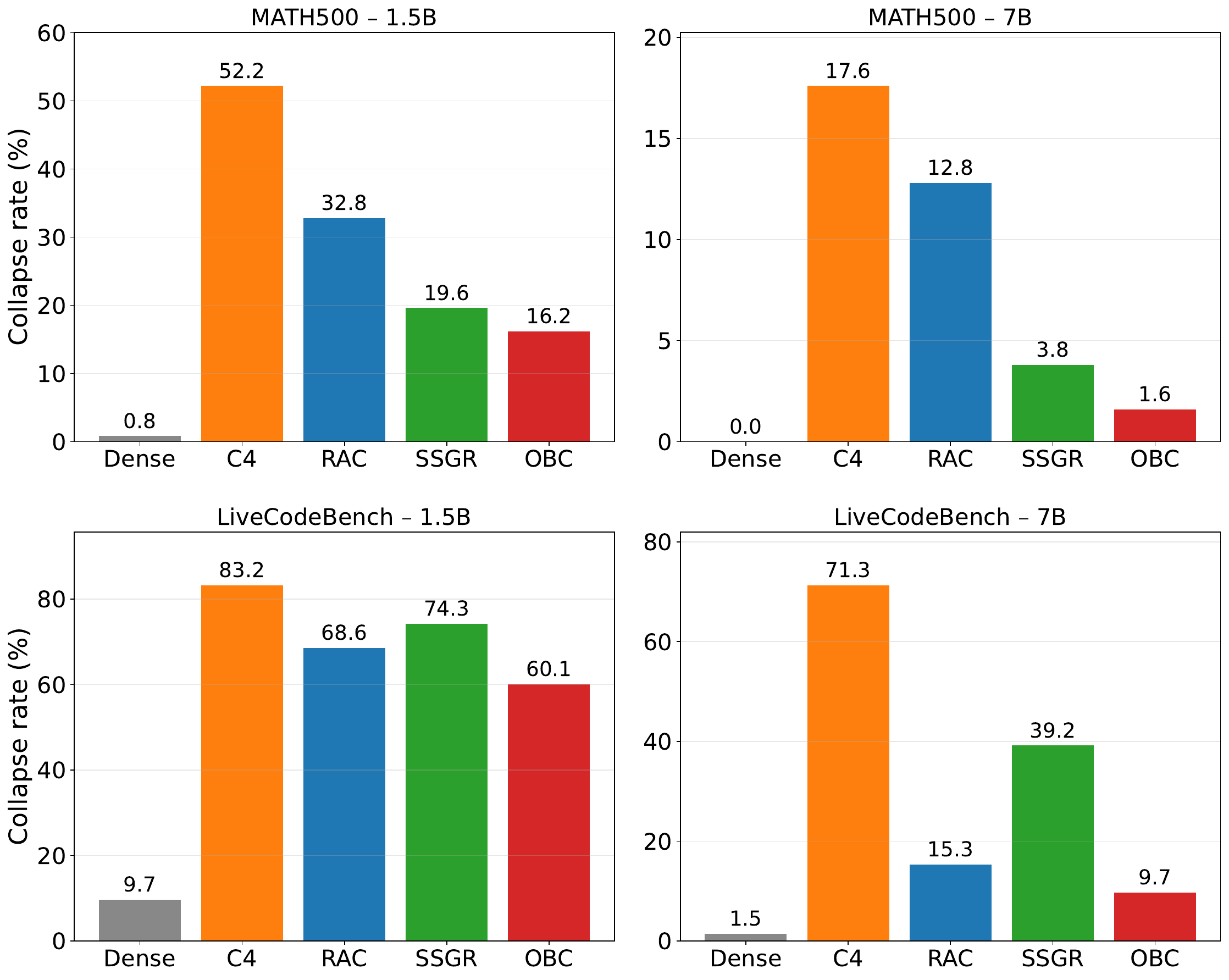}
\caption{Collapse rate (completions reaching $\geq$98\% of the observed max generation length) under C4, RAC, SSGR, and OBC-Prune calibration using SparseGPT, for the DeepSeek-R1-Distill-Qwen 1.5B and 7B models on MATH500 and LiveCodeBench.}
\label{fig:collapse_rate}
\end{figure}

\paragraph{Collapse rate (RQ3).} Figure~\ref{fig:collapse_rate} reports the fraction of completions reaching $\geq$98\% of the observed max generation length at 50\% sparsity, a proxy for the model failing to terminate its chain of thought, treated here as a reasoning failure distinct from an incorrect final answer. On MATH500, where dense collapse is negligible (0.8\% at 1.5B, 0.0\% at 7B), collapse rises sharply under pruning but by very different amounts across methods: C4 collapses most (52.2\%/17.6\%), followed by RAC (32.8\%/12.8\%) and SSGR (19.6\%/3.8\%), with OBC-Prune lowest at both scales (16.2\%/1.6\%). Regarding LiveCodeBench the collapse rate of OBC-Prune remains lowest for both model sizes. This is consistent with our reconstruction-fidelity results: OBC-Prune-pruned models retain the ability to recognize when reasoning is complete better than other methods.

\begin{figure}[h]
\centering
\includegraphics[width=1.0\linewidth]{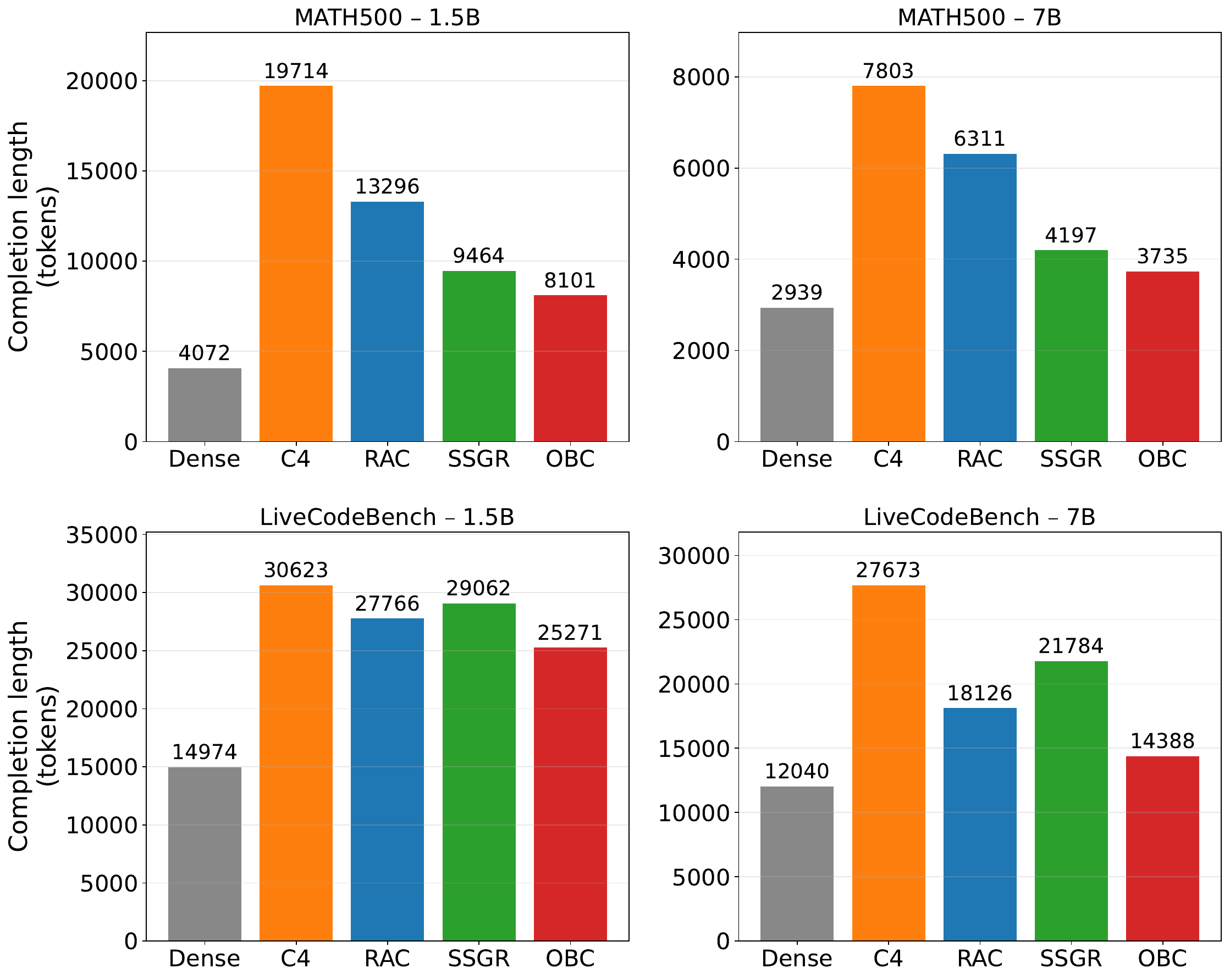}
\caption{Mean completion length (tokens) at 50\% sparsity under C4, RAC, SSGR, and OBC-Prune calibration with SparseGPT for the DeepSeek-R1-Distill-Qwen 1.5B and 7B models on MATH500 and LiveCodeBench.}
\label{fig:completion_length}
\end{figure}

\paragraph{Completion length (RQ3).} Figure~\ref{fig:completion_length} shows mean completion length at 50\% sparsity, which together with collapse rate reflects whether pruning disrupts the model's stopping capability. On MATH500, OBC-Prune stays closest to dense at both scales (1.5B: 8{,}101 vs.\ 4{,}072 dense, versus 13{,}296 for RAC and 19{,}714 for C4; 7B: 3{,}735 vs.\ 2{,}939 dense, versus 6{,}311 for RAC and 7{,}803 for C4), roughly $1.6$--$2\times$ shorter than the baselines. The same ordering holds on LiveCodeBench, though the gap narrows in relative terms (1.5B: 25{,}271 vs.\ 27{,}766--30{,}623 for baselines; 7B: 14{,}388 vs.\ 18{,}126--27{,}673). This indicates baseline-calibrated models tend to ramble longer under pruning, consuming more inference budget per query, whereas OBC-Prune keeps completions closest to the dense model's length across both benchmarks and model sizes.


\paragraph{Generalization to other model families (RQ4).} To test whether OBC-Prune's advantage is specific to the DeepSeek-R1-Distill-Qwen family, we repeated the SparseGPT/40\%/MATH500 comparison on two additional model families: Qwen3 (1.7B/8B/14B) and DeepSeek-R1-Distill-Llama-8B, a reasoning model distilled onto a different base architecture. Appendix~\ref{sec:cross_family_appendix} shows OBC-Prune matches or exceeds C4, RAC, and SSGR at every model and sparsity, with the larger gains at the smaller model and higher sparsity. This mirrors the scale-dependent pattern already observed for DeepSeek-R1-Distill-Qwen (Table~\ref{tab:pruning_results}), where smaller models, with less redundant capacity, benefit more from calibration that preserves causally-important reasoning circuits.

\paragraph{Sparsity sweep to collapse (RQ5).} Figure~\ref{fig:sparsity_sweep} sweeps sparsity from 10\% to 90\% on DeepSeek-R1-Distill-Qwen (1.5B/7B) to locate where each method's accuracy collapses below a 5\% threshold. All four methods track closely up to 30\% sparsity; OBC-Prune's advantage then opens up from 40\% onward, peaking at 50--60\%, before all methods converge near zero past 70\%. At 1.5B, C4, RAC, and SSGR drop below threshold between 60--70\% sparsity, while OBC-Prune holds on until 80\% (roughly 6\% accuracy at 70\%). At 7B, the same one-step delay appears higher up: baselines collapse between 70--80\%, while OBC-Prune stays well above threshold at 70\% (roughly 46\% accuracy) before collapsing at 80\%. This indicates the calibration objective delays, but does not eliminate, collapse under aggressive one-shot pruning.

\paragraph{Hyperparameter sensitivity (RQ5).} To check that OBC-Prune's gains are not an artifact of hand-tuned $\beta$, $\gamma$, $\varepsilon$, we swept each hyperparameter one-at-a-time (holding the other two at their paper defaults $\beta{=}2$, $\gamma{=}0.5$, $\varepsilon{=}0.1$), using the identical calibration pairs to prune the model DeepSeek-R1-Distill-Qwen 1.5B at 40\% sparsity. Appendix~\ref{sec:hparam_sweep} shows accuracy is stable across all settings.

\section{Conclusion}
\label{sec:conclusion}

We presented OBC-Prune, a calibration framework for pruning reasoning language models that stratifies on-policy chain-of-thought calibration data by outcome. By pairing correct and incorrect rollouts on difficulty-matched problems and scoring each reasoning sentence's causal contribution through an attention-suppression intervention, OBC-Prune upweights the activations behind successful reasoning and downweights those behind reasoning failures, without modifying the underlying pruning solver. Across DeepSeek-R1-Distill-Qwen at 1.5B, 7B, and 14B, OBC-Prune matches or improves on C4, RAC, and SSGR calibration in most model/sparsity settings on MATH500, LiveCodeBench, and AIME 2025, with gains that generalize across three structurally distinct pruning backends (SparseGPT, Wanda, and ALPS). OBC-Prune's pruned models track the dense model's internal computation more faithfully, and terminate their reasoning more reliably. This advantage extends beyond the DeepSeek-R1-Distill-Qwen family to other reasoning model families, and is robust to the choice of hyperparameters. Together, these results support our central claim: preserving the specific circuits that are causally responsible for correct reasoning is a more effective pruning objective than preserving all activations uniformly. More broadly, our findings suggest that the composition of the calibration set is as important as the pruning criterion, and that outcome-based calibration can improve pruning quality without additional training.

\bibliography{aaai2027}


\newpage

\appendix

\section{Calibration-Statistic Re-Weighting for Wanda and ALPS}
\label{sec:backend_reweighting}

Section~\ref{sec:methodology} (Stage 4) derives OBC-Prune's calibration re-weighting for
SparseGPT. Here we give the analogous derivation for the backends used in
Section~\ref{sec:experiments} (RQ2).

\begin{itemize}
  \item \textbf{Wanda} \cite{wanda} accumulates a per-channel activation-energy statistic
        $\sum_t \|x_t\|_2^2$ with no cross terms and performs no compensation update;
        rescaling gives $\sum_t w_t \|x_t\|_2^2$ under the identical $\sqrt{w_t}$ token
        scaling, leaving Wanda's pruning score and lack of weight update otherwise unchanged.
  \item \textbf{ALPS} \cite{alps} accumulates the same Gram matrix
        $\mathbf{X}_\ell \mathbf{X}_\ell^\top$ as SparseGPT, solved via ADMM followed by
        conjugate-gradient polishing rather than SparseGPT's greedy OBS updates; it is
        re-weighted identically to SparseGPT's Hessian above, with its ADMM solver applied
        unmodified to $\tilde{\mathbf{H}}_\ell$ in place of $\mathbf{H}_\ell$.
\end{itemize}

\section{Additional Backend Result (RQ2)}
\label{sec:alps_appendix}
\label{sec:alps}
\begin{table*}[h]
\centering
\small
\begin{tabular}{llcccccccc}
\toprule
\multirow{2}{*}{Model} & \multirow{2}{*}{Sparsity} &
\multicolumn{4}{c}{Accuracy $\pm$ SE} &
\multicolumn{4}{c}{Runtime (min)} \\
\cmidrule(lr){3-6}
\cmidrule(lr){7-10}
& & C4 & RAC & SSGR & OBC-Prune 
& C4 & RAC & SSGR & OBC-Prune \\
\midrule

\multirow{3}{*}{1.5B}
& Dense & \multicolumn{4}{c}{0.802 (0.018)} & \multicolumn{4}{c}{6.8} \\
& 40\%
& 0.672 (0.021)
& 0.698 (0.021)
& 0.778 (0.019)
& \textbf{0.780 (0.019)}
& 18.50
& 15.17
& 14.40& \textbf{9.72 }\\
& 50\%
& 0.398 (0.022)
& 0.574 (0.022)
& 0.732 (0.020)
& \textbf{0.754 (0.019)}
& 39.30
& 20.98
& 18.20& \textbf{12.60} \\

\cmidrule(lr){1-10}

\multirow{3}{*}{7B}
& Dense & \multicolumn{4}{c}{0.886 (0.014)} & \multicolumn{4}{c}{8.5} \\
& 40\%
& 0.844 (0.016)
& 0.852 (0.016)
& 0.858 (0.014)& \textbf{0.870 (0.015)}
& 10.50
& 9.28
& 8.50& \textbf{7.82} \\
& 50\%
& 0.760 (0.019)
& 0.776 (0.019)
& 0.852 (0.016)
& \textbf{0.858 (0.016)}
& 20.60
& 16.75
& 11.50& \textbf{11.12} \\

\cmidrule(lr){1-10}

\multirow{3}{*}{14B}
& Dense & \multicolumn{4}{c}{0.914 (0.014)} & \multicolumn{4}{c}{14.4} \\
& 40\%
& 0.880 (0.015)
& 0.892 (0.014)
& 0.896 (0.014)
& \textbf{0.900 (0.013)}
& 25.90
& 16.02
& 16.60
& \textbf{15.43} \\
& 50\%
& 0.760 (0.019)
& 0.864 (0.015)
& 0.864 (0.014)& \textbf{0.880 (0.015)}
& 44.60
& 31.25
& 21.90
& \textbf{17.30} \\

\bottomrule
\end{tabular}
\caption{Accuracy with standard error (SE) and runtime comparison for ALPS-pruned DeepSeek-R1-Distill-Qwen models (1.5B/7B/14B) using different calibration datasets (C4, RAC, SSGR, and OBC-Prune) at 40\% and 50\% sparsity on MATH500~\cite{math500_hf} (acc@1). Dense (unpruned) accuracy and evaluation runtime are shown for reference.}
\label{tab:alps_results}
\end{table*}
Table~\ref{tab:alps_results} reports the ALPS counterpart to Table~\ref{tab:wanda_results}.
OBC-Prune remains the best-performing calibration method in every model and sparsity
setting, in both accuracy and runtime, extending the pattern already established under
SparseGPT and Wanda to a third, structurally distinct backend. As with the other two
backends, OBC-Prune's advantage over the strongest baseline widens as sparsity increases,
reinforcing RQ1's central finding that its gains grow with sparsity rather than being
confined to a single operating point. This holds even though ALPS's ADMM-based
compensation update makes it the least calibration-sensitive of the three backends
(Section~\ref{sec:preliminaries}). Runtime tracks the same
ordering, with OBC-Prune the fastest calibration method throughout and the gap over other baselines widening at larger sparsity.

\section{Generalization across model families and architectures (RQ4)}
\label{sec:cross_family_appendix}

\begin{table*}[h]
\centering
\small
\caption{Accuracy and runtime comparison of different calibration strategies for SparseGPT across additional model families and sparsity levels.}
\begin{tabular}{llcccccccc}
\toprule
\multirow{2}{*}{Model} &
\multirow{2}{*}{Sparsity} &
\multicolumn{4}{c}{Accuracy $\pm$ SE} &
\multicolumn{4}{c}{Runtime (min)} \\
\cmidrule(lr){3-6}\cmidrule(lr){7-10}
\setlength{\tabcolsep}{3pt}
&& C4 & RAC & SSGR & OBC-Prune & C4 & RAC & SSGR & OBC-Prune \\
\midrule
\multirow{3}{*}{Qwen3 1.7B}
& Dense & \multicolumn{4}{c}{0.884 (0.014)} & \multicolumn{4}{c}{14.0} \\
& 40\% & 0.330 (0.021) & 0.708 (0.020) & 0.806 (0.018) & \textbf{0.832 (0.017)} & 146.3 & 53.6 & 23.5 & \textbf{19.4} \\
& 50\% & 0.048 (0.010) & 0.370 (0.022) & 0.520 (0.022) & \textbf{0.548 (0.022)} & 215.0 & 128.2 & 62.9& \textbf{50.7} \\

\addlinespace

\multirow{3}{*}{Qwen3 8B}
& Dense & \multicolumn{4}{c}{0.948 (0.010)} & \multicolumn{4}{c}{21.5} \\
& 40\% & 0.870 (0.015) & 0.934 (0.011) & 0.944 (0.009) & \textbf{0.946 (0.010)} & 31.3 & 39.3 & 22.5 & \textbf{22.2} \\
& 50\% & 0.548 (0.022) & 0.812 (0.018) & 0.934 (0.010) & \textbf{0.944 (0.010)} & 185.6 & 137.4 & 138.5 & \textbf{35.1} \\

\addlinespace

\multirow{3}{*}{Qwen3 14B}
& Dense & \multicolumn{4}{c}{0.960 (0.015)} & \multicolumn{4}{c}{22.0} \\
& 40\% & 0.840 (0.016) & 0.906 (0.013) & 0.888 (0.014) & \textbf{0.948 (0.012)}& 33.1 & 30.8 & 27.9 & \textbf{25.0} \\
& 50\% & 0.812 (0.018) & 0.834 (0.017) & 0.876 (0.015) & \textbf{0.944 (0.013)}& 52.7 & 74.8 & 31.6 & \textbf{26.1} \\

\addlinespace

\multirow{3}{*}{DS-R1-Llama 8B}
& Dense & \multicolumn{4}{c}{0.824 (0.017)} & \multicolumn{4}{c}{11.5} \\
& 40\% & 0.688 (0.021) & 0.696 (0.021) & 0.762 (0.019) & \textbf{0.764 (0.019)} & 19.2 & 22.3 & 18.9 & \textbf{17.6} \\
& 50\% & 0.492 (0.022) & 0.538 (0.022) & \textbf{0.670 (0.021)} & \textbf{0.670 (0.021)} & 66.9 & 53.7 & 30.5 & \textbf{28.7} \\
\bottomrule
\end{tabular}
\label{tab:cross_family}
\end{table*}

Table 4 shows OBC-Prune's advantage is not specific to the DeepSeek-R1-Distill-Qwen family. The improvement over baselines is most pronounced for the smallest model at the highest sparsity level. At 1.7B parameters with 50\% sparsity, calibrating on C4 (a generic corpus) causes accuracy to collapse to near-random levels (0.048). RAC and SSGR partially recover performance (0.370 and 0.520, respectively), while OBC-Prune achieves the best result (0.548). This suggests that generic-corpus calibration is least reliable precisely when redundant model capacity is scarcest, i.e., in small models pruned aggressively. While OBC-Prune's advantage over the weakest baseline (C4) shrinks at larger scale, its margin over the strongest baseline (SSGR) does not: at 14B/50\%, OBC-Prune still leads SSGR by 0.068 (0.944 vs. 0.876), a larger absolute gap than at 1.7B (0.028), indicating that the benefit of outcome-based calibration is not confined to small, capacity-constrained models. DeepSeek-R1-Distill-Llama-8B, which pairs the same distillation recipe with a different base architecture, shows the same ordering as its Qwen counterpart at comparable scale, with OBC-Prune matching SSGR at 50\% sparsity (0.670), suggesting the residual gap between OBC-Prune and SSGR narrows as the calibration budget becomes the binding constraint rather than the calibration objective itself. The same pattern holds for runtime, with the largest gap at Qwen3-8B at 50\% sparsity, where it is nearly $4\times$ faster than SSGR (35.1 vs.\ 138.5 minutes) while achieving comparable accuracy. This indicates that OBC-Prune achieves similar calibration quality at a much lower runtime cost than other methods.

\section{Accuracy and sparsity sweep (RQ5)}
\begin{figure}[h]
\centering
\includegraphics[width=1.0\linewidth]{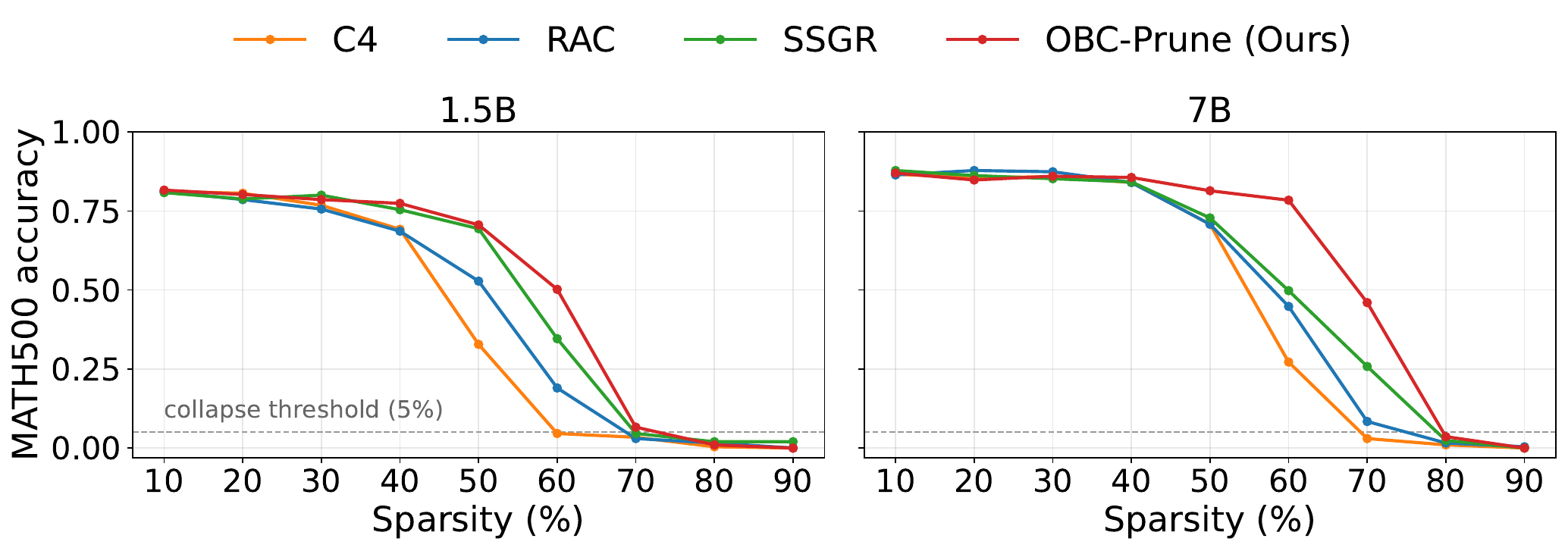}
\caption{MATH500 accuracy under C4, RAC, SSGR, and OBC-Prune (SparseGPT) as sparsity is
swept from 10\% to 90\% in 10-point increments, for DeepSeek-R1-Distill-Qwen models
(1.5B/7B). Dashed horizontal line marks the 5\% collapse threshold.}
\label{fig:sparsity_sweep}
\end{figure}

The sweep prunes DeepSeek-R1-Distill-Qwen 1.5B and 7B at each sparsity level from 10\%
to 90\% in 10-point increments under SparseGPT, and evaluates on MATH500 with the
generation settings described in Section~\ref{sec:experiments}. 
All four calibration methods (C4, RAC, SSGR, and OBC-Prune) reuse the same calibration construction as in the main experiments at each sparsity level. The OBC-Prune hyperparameters ($\beta{=}2$, $\gamma{=}0.5$, $\varepsilon{=}0.1$) are held fixed across all nine sparsity levels rather than re-tuned, ensuring that any widening or narrowing of the performance gap reflects each calibration strategy's robustness to increasing sparsity rather than differences in per-level hyperparameter optimization. We conduct this sweep on MATH500 because, among our benchmarks, it contains the largest number of problems, yielding the smallest standard errors and thus the greatest statistical power for resolving how accuracy changes with sparsity and where the performance of different calibration methods begins to diverge.

\section{Hyperparameter sensitivity (RQ5)}
\label{sec:hparam_sweep}
\begin{table}[h]
\centering
\small
\begin{tabular}{clc}
\toprule
Hyperparameter & Value & MATH500 acc@1 \\
\midrule
Default ($\beta{=}2$, $\gamma{=}0.5$, $\varepsilon{=}0.1$) & — & 0.774 (0.019) \\
\midrule
\multirow{2}{*}{$\beta$} & 1.0 & 0.766 (0.019) \\
                          & 3.0 & 0.774 (0.019) \\
\midrule
\multirow{2}{*}{$\gamma$} & 0.25 & 0.770 (0.020) \\
                           & 0.75 & 0.778 (0.019) \\
\midrule
\multirow{2}{*}{$\varepsilon$} & 0.05 & 0.770 (0.019) \\
                                & 0.2  & 0.770 (0.019) \\
\bottomrule
\end{tabular}
\caption{Hyperparameter sensitivity sweep (1.5B, SparseGPT, 40\% sparsity, MATH500 acc@1 $\pm$ SE). Each row sweeps one hyperparameter with the other two held at their default (top row).}
\label{tab:hparam_sweep}
\end{table}

Table~\ref{tab:hparam_sweep} sweeps each hyperparameter individually around its default setting ($\beta{=}2$, $\gamma{=}0.5$, $\varepsilon{=}0.1$) on the Deepseek-R1-Distill-Qwen 1.5B model at 40\% sparsity. Across all settings, accuracy remains within one standard error of the default, indicating that OBC-Prune is robust to moderate variations in its hyperparameters. Varying the correct-rollout upweighting factor $\beta$ produces only minor changes in accuracy, while varying the wrong-rollout downweighting factor $\gamma$ leads to similarly small fluctuations. Likewise, changing the minimum wrong-rollout weight $\varepsilon$ over a fourfold range (0.05--0.2) has little effect on performance. Overall, these results suggest that OBC-Prune does not rely on precise hyperparameter tuning and achieves stable performance across a reasonable range of hyperparameter values.

\section{Component Ablation Study}
\label{sec:component_ablation}

Appendix~\ref{sec:hparam_sweep} shows that OBC-Prune is not sensitive to the exact choice of $\beta$, $\gamma$, and $\varepsilon$. Here we ask a different question: which \emph{components} of
the re-weighting scheme drive its accuracy gains, as opposed to merely the specific
constants chosen for them. Starting from the same calibration pairs, sentence segments,
and causal-effect scores used for the paper's main $\beta{=}2$, $\gamma{=}0.5$,
$\varepsilon{=}0.1$ run, we disable one component at a time and re-prune
DeepSeek-R1-Distill-Qwen-1.5B to 40\% sparsity with SparseGPT, evaluating on MATH500 acc@1:

\begin{itemize}
  \item \textbf{No wrong-rollout downweighting} ($\gamma{=}0$): correct-rollout tokens are
        still upweighted by their causal-effect score, but wrong-rollout tokens revert to
        a uniform weight of $1$ -- isolating the contribution of downweighting reasoning
        that led to an incorrect answer.
  \item \textbf{No correct-rollout upweighting} ($\beta{=}0$): wrong-rollout tokens are
        still downweighted, but correct-rollout tokens revert to a uniform weight of $1$
        -- isolating the contribution of upweighting reasoning that led to a correct
        answer.
  \item \textbf{No softmax normalisation}: sentence scores are min-max, rather than
        softmax, normalised across the rollout (Section~\ref{sec:stage2}) before
        $\beta$/$\gamma$ are applied, replacing softmax's exponential compression with a
        linear rescaling of the baseline-subtracted causal effects.
  \item \textbf{Uniform span weights}: the causal-effect score is discarded entirely and
        every sentence in a rollout is assigned an equal weight $\tilde{e}(k,R) = 1/M$ ($M$ is the number of sentence spans in the rollout, as segmented in Section~\ref{sec:stage2}), so only the coarse correct/wrong split (via $\beta$/$\gamma$) determines the per-token weight, without considering which sentences are more or less causally important.
\end{itemize}

\begin{table}[h]
\centering
\small
\begin{tabular}{lc}
\toprule
Ablation & MATH500 acc@1 \\
\midrule
Full OBC-Prune ($\beta{=}2$, $\gamma{=}0.5$, softmax) & \textbf{0.774 (0.019)} \\
\midrule
No wrong-rollout downweighting ($\gamma{=}0$) & 0.762 (0.019) \\
No correct-rollout upweighting ($\beta{=}0$) & 0.754 (0.019) \\
No softmax normalisation (min-max) & 0.746 (0.019) \\
Uniform span weights & 0.750 (0.019) \\
\bottomrule
\end{tabular}

\caption{Component ablation of OBC-Prune's re-weighting scheme (1.5B, SparseGPT, 40\%
sparsity, MATH500 acc@1 $\pm$ SE). Each row disables exactly one component while holding
the calibration pairs and causal-effect scores fixed.}
\label{tab:component_ablation}
\end{table}
Every ablation underperforms full OBC-Prune, confirming that no single component is
redundant, but the components contribute unevenly. Removing wrong-rollout downweighting
alone costs the least (0.762 vs.\ 0.774), whereas removing correct-rollout upweighting
costs more (0.754), suggesting that upweighting causally important correct reasoning
matters somewhat more than downweighting causally important wrong reasoning at this
sparsity. Replacing the softmax normalization with min-max normalization (0.746) or discarding the causal-effect scores entirely by assigning uniform span weights (0.750) produces the largest drops in accuracy. 

This suggests that the coarse correct/wrong split alone is insufficient. Without a sentence-level causal signal identifying which parts of each reasoning trace should be emphasized or suppressed, much of OBC-Prune's advantage is lost. Together, these results indicate that sentence-level causal weighting, rather than the correct/incorrect split alone, is a key contributor to OBC-Prune's performance gains.
\end{document}